\documentclass[conference]{IEEEtran}
\ifCLASSINFOpdf
\else
\fi
\usepackage[english]{babel}
\usepackage[utf8]{inputenc}

\usepackage{mathrsfs}
\usepackage{graphicx}
\usepackage{subfigure}
\usepackage{pdfpages}
\usepackage{amsmath}

\usepackage{algorithm}
\usepackage{algcompatible}

\begin{document}
%
\title{Evaluating Competence Measures for Dynamic Regressor Selection}

\author{

\IEEEauthorblockN{Thiago J. M. Moura}
\IEEEauthorblockA{Centro de Informática (CIn)\\Universidade Federal de Pernambuco\\ Recife - PE, Brazil\\Instituto Federal da Paraíba (IFPB)\\
João Pessoa - PB, Brazil\\
Email: tmoura@gmail.com}
\and
\IEEEauthorblockN{George D. C. Cavalcanti}
\IEEEauthorblockA{Centro de Informática (CIn)\\Universidade Federal de Pernambuco\\Recife - PE, Brazil\\
Email: gdcc@cin.ufpe.br}
\and
\IEEEauthorblockN{Luiz S. Oliveira}
\IEEEauthorblockA{Departamento de Informática (DInf)\\Universidade Federal do Paraná\\ Curitiba - PR, Brazil\\Email: luiz.oliveira@ufpr.br}}


%


\maketitle

\begin{abstract}
Dynamic regressor selection (DRS) systems work by selecting the most competent regressors from an ensemble to estimate the target value of a given test pattern. This competence is usually quantified using the performance of the regressors in local regions of the feature space around the test pattern. However, choosing the best measure to calculate the level of competence correctly is not straightforward. The literature of dynamic classifier selection presents a wide variety of competence measures, which cannot be used or adapted for DRS.
In this paper, we review eight measures used with regression problems, and adapt them to test the performance of the DRS algorithms found in the literature. Such measures are extracted from a local region of the feature space around the test pattern, called region of competence, therefore competence measures.
To better compare the competence measures, we perform a set of comprehensive experiments of 15 regression datasets. Three DRS systems were compared against individual regressor and static systems that use the Mean and the Median to combine the outputs of the regressors from the ensemble. The DRS systems were assessed varying the competence measures. Our results show that DRS systems outperform individual regressors and static systems but the choice of the competence measure is problem-dependent.

\end{abstract}


%
\IEEEpeerreviewmaketitle

\section{Introduction}

Model selection systems consist of two main phases \cite{Cruz2018}: Generation and Selection. In the first phase, a training set is used to generate the ensemble. The ensemble is said homogeneous when a single learning algorithm is used to train the models. Otherwise, it is called heterogeneous. In the second phase, a model or a subset of models are selected to evaluate the test pattern. Such a selection can be done according to two distinct approaches: static and dynamic. In the static approach, selection occurs using the performance of the models in the training set \cite{OrtizBoyer2005} or using a separated validation set after the training stage \cite{Partalas2008}. In the static selection, the models are used to evaluate all test patterns. In the dynamic approach, a different model or subset of models from the ensemble are selected for each new test pattern. In dynamic selection techniques, each model is an expert in a specific local region of the feature space, which is known as region of competence. So, for each test pattern, the most competent models are selected for the region of competence where the test pattern is located. The region of competence contains the patterns from the training set or the validation set which are neighbors of the test pattern, also known as the neighborhood of the test pattern. The standard method for defining the region of competence is the  k-nearest neighbors ($k$NN) algorithm with Euclidean distance \cite{Woods1997}. Dynamic selection is a growing research area in machine learning, and recent works have shown that dynamic selection techniques are more efficient than static selection for both classification and regression problems \cite{Britto2014,survey}.

\begin{figure*}[!t]
\centering
\includegraphics{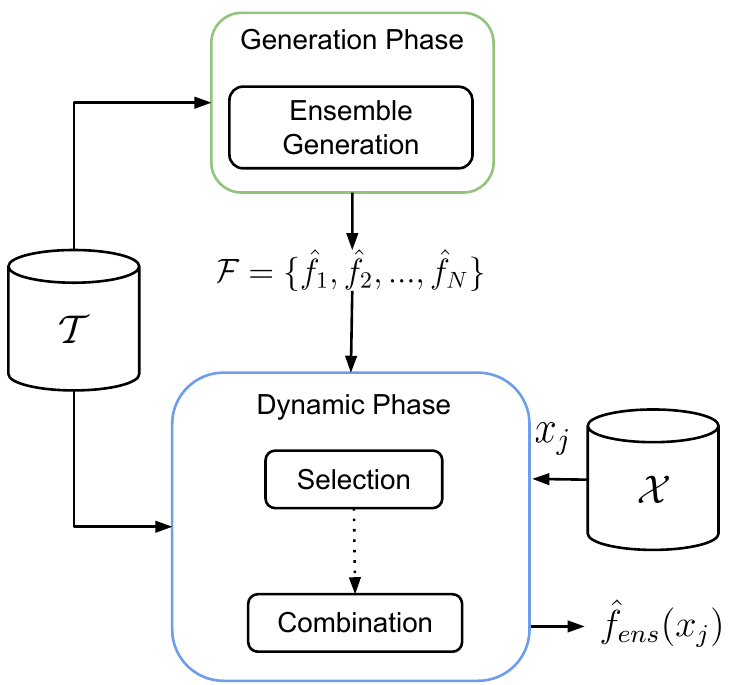}
\caption{Overview of the Dynamic Regressor Selection architecture. $\mathcal{T}$ and $\mathcal{X}$ are the training and testing sets respectively. $\mathcal{F}$ is the ensemble of regressors generated in the Generation Phase, $x_j$ is a test pattern and $\hat{f}_{ens}(x_j)$ is the result of the test pattern estimate.}
\label{fig:arqGeral}
\end{figure*}

The central issue in dynamic selection is to define the criterion to measure the competence of each model in the ensemble, i.e., the competence measure. In general, the accuracy of the models in the region of competence is used as a criterion for determining the competence. Some works of dynamic classifier selection (DCS) \cite{Santana2006}, \cite{DosSantos2008} use other measures, beyond accuracy, to calculate the competence. Recent works on DCS \cite{Cruz2015a}, \cite{Cruz2015}, \cite{Cruz2017} use the composition of many measures to determine the competence of the classifiers, selecting and combining them to predict the class of the test pattern. 

Many of the measures used in DCS cannot be directly used for regression. So, dynamic regressor selection (DRS) literature methods commonly use the error of the predictions in the region of competence as a criterion to dynamically select the best regressors, e.g., Rooney et al.~\cite{Rooney2004} and Moreira et al.~\cite{Mendes-Moreira2009}. Rooney et al. adapted the DCS technique proposed by Tsymbal et al. \cite{Tsymbal2000}, \cite{Tsymbal2006} for regression problems by defining three DRS algorithms: Dynamic Selection (DS), Dynamic Weighting (DW), and Dynamic Weighting with Selection (DWS). The algorithms dynamically select the most competent regressors in the region of competence per test pattern. Such algorithms use the performance of the regressors in the region of competence as a selection criterion; it means that the regressors with the smallest cumulative error in the neighborhood are chosen to estimate the test pattern. Moreira et al. \cite{Mendes-Moreira2009} also use the error similarly to Rooney et al., however, in their work, the estimated errors are weighted by the distance between the patterns in the region of competence and the test pattern.

Having in mind that literature uses the prediction error in the region of competence as a criterion to select the best regressors, we assume that DRS algorithms can benefit from different criteria to select the best regressors per query pattern. So, we perform an empirical evaluation of eight measures that may employed as a criterion to measure the competence (competence measures) of the regressors for DRS. 

To the best of our knowledge, seven of these measures are adapted for the first time to this task, and they capture different information, such as weighted error, variance, and similarity. These eight competence measures are evaluated using 15 regression problems from different data repositories and three literature algorithms: DS, DW, and DWS.

The contributions of this work are: i) Evaluation of eight competence measures that are used as a criterion to select the best regressors per query pattern. Seven of these eight measures are evaluated for the first time in this task; ii) Comparative study using three dynamic selection algorithms (DS, DW, and DWS) and all the competence measures; iii) Comparison between dynamic systems and individual regressor; iv) Comparison between dynamic and static systems that use the Mean and the Median to combine the outputs of the regressors from the ensemble. 


This paper is organized as follows. Section \ref{sec:arquitetura} presents the dynamic regressors selection algorithms. Section~\ref{sec:caracteristicas} describes the eight measures. The experimental results are shown in Section~\ref{sec:experimentos} and the final remarks are presented in Section \ref{sec:conclusoes}.

\section{DRS algorithms}
\label{sec:arquitetura}

A general overview of the DRS architecture is depicted in Figure \ref{fig:arqGeral}. The architecture is divided into two phases: Generation and Dynamic. They are described in the following subsections.

\subsection{Generation Phase}

This phase generates an ensemble $\mathcal{F} = \{\hat{f}_1, \hat{f}_2,...,\hat{f}_N\}$ using the training set $\mathcal{T}$, where $N$ is the number of regressors. The ensemble can be homogeneous or heterogeneous. The former uses different sets to train each regressor $\hat{f}_n, \forall n \in \{1,2,...,N\}$. These different sets are commonly generated using Bagging \cite{bagging}, Boosting \cite{Shrestha2006}, or Random Subspace \cite{Ho1998}. Heterogeneous ensembles, on the other hand, are generated using different learning algorithms for training the regressors.

\subsection{Dynamic Phase}
\label{sec:ag}

The Dynamic phase selects a subset of the regressors per test pattern $x_j \in \mathcal{X}$ and it can work in three different ways: (I) select only one regressor from the ensemble $\mathcal{F}$; (II) weighted combination of all the regressors and; (III) select a subset of the regressors and combine them. The result of this phase is the ensemble prediction $\hat{f}_{ens}(x_j)$.

Any of the three algorithms proposed in~\cite{Rooney2004} for the Dynamic phase can be used in the architecture depicted in Figure~\ref{fig:arqGeral}. All of them use the concept of region of competence. Given a test pattern $x_j$, its region of competence is a set $\Psi$ composed of the $K$ nearest neighbours of $x_j$ in the validation or training set given by $\{t_1, t_2,...,t_K\}$. These three algorithms were described by Moreira et al.~\cite{Mendes-Moreira2009,Mendes-Moreira2015} as follows:

\begin{itemize}

\item Dynamic Selection (DS) - it selects the regressor with the lowest accumulated error in the region of competence. The errors are weighted by the distance between the neighborhood pattern and the test pattern. Only a single regressor is selected and no combination is required. The estimation of the test pattern is the value returned by the selected regressor.

\item Dynamic Weighting (DW) - it combines all the regressors of the ensemble using the weighted mean. For each test pattern $x_j$, its region of competence $\Psi$ is calculated; $\Psi$ is composed of $K$ patterns. For each pattern in $\Psi$, a weight is calculated using Equation~\ref{eq:vectorWeights}:

\begin{equation}
\label{eq:vectorWeights}
d_k = \frac{\frac{1}{dist_k}}{\sum_{j=1}^{K}(\frac{1}{dist_{j}})}
\end{equation} 

\noindent where $dist_k$ is a distance measure between a pattern $t_k \in \Psi$ and the test pattern $x_j$. 

The vector $\{d_1,d_2,...,d_k\}, k \in \{1,2,...,K\}$, is used to calculate the weight $\alpha_i$ of the regressor $\hat{f}_i$ using Equation~\ref{eq:dwAlgorithm}: 


\begin{equation}
\label{eq:dwAlgorithm}
\alpha_i = \frac{\frac{1}{\sqrt{\sum_{k=1}^{K}(d_k \times sqe_{k,i})}}}{\sum_{n=1}^{N}\left(\frac{1}{\sqrt{\sum_{k=1}^{K}(d_k \times sqe_{k,n})}}\right)}
\end{equation} 

\noindent where $N$ is the ensemble size, $k$ represents the index of the neighbor, and $sqe_{k,i}$ is the squared error of the regressor $i$ calculated using the pattern $t_k \in \Psi$.

\item Dynamic Weighting with Selection (DWS) - it combines a subset of the regressors. The regressors with the accumulated error in the upper half of the error interval $E_i > (E_{max} - E_{min})/2$ are discarded, where $E_{max}$ is the largest accumulated error of any regressor and $E_{min}$ is the lowest accumulated error of any regressor. The measure to calculate the performance of the regressors from the ensemble is the same than the DW algorithm and the remaining regressors are combined using the same strategy of the DW.
\end{itemize}

In \cite{Mendes-Moreira2009}, the authors use the Root Sum Squared Error as a competence measure to select a regressor from the ensemble for the three aforementioned algorithms. In Section \ref{sec:caracteristicas}, this competence measure is defined as $m_7$. In spite of the good results reported in \cite{Mendes-Moreira2009}, we show that other competence measures perform better and should not be neglected. To the best of our knowledge, this is the first work that analyzes other competence measures for DRS algorithms.

\section{Competence Measures}
\label{sec:caracteristicas}

Table~\ref{tab:measures} shows a summary of all competence measures used in this work. These measures correspond to different criteria to measure the behavior of each regressor from the ensemble $\mathcal{F}$. Each measure expresses one of the following information: (i) the error calculated in the region of competence; (ii) the variance of the estimated values in the neighborhood; or (iii) the similarity between the observed and the estimated values of the test pattern $x_j$.

Some competence measures are calculated using the distances between the test pattern and the nearest neighbors. However, instead of using the distance, we use the inverse of the normalized distance ($d_k$) in the interval [0,1]. So, smaller the distance greater the value of $d_k$, according to Equation \ref{eq:vectorWeights}.

The measures described below are extracted using the region of competence $\{t_1, t_2,...,t_K\}$ of the test pattern $x_j$. In the next equations, $f(t_k)$ stands for the observed value of the neighborhood pattern and $\hat{f}_n(t_k)$ is the estimated value of the pattern $t_k$ given by the regressor $\hat{f}_n$.

\begin{itemize}

\item $m_1$ - \textit{Variance}: the variance of the neighbors estimated values. The variance is calculated for each regressor using the estimated values of the patterns in the region of competence, according to Equation \ref{eq:variance}: \begin{equation}
\label{eq:variance}
m_1 = Var({\hat{f}_n(t_1),\hat{f}_n(t_2),...,\hat{f}_n(t_K)})
\end{equation} 

This competence measure is inspired in the work of Tresp et al. \cite{Tresp1995}, whose variance of the estimated values is used as weight in the static combination of artificial neural networks.

\item $m_2$ - \textit{Sum Absolute Error}: the sum of the absolute errors is calculated in the region of competence, weighted by $d_k$, according to Equation \ref{eq:MAE}: \begin{equation}
\label{eq:MAE}
m_2 = \sum_{k=1}^{K}\left|f(t_k) - \hat{f}_n(t_k)\right| \times d_k
\end{equation}

\item $m_3$ - \textit{Sum Squared Error}: the sum of the squared errors is calculated using the inverse of the distances $d_k$ as weights, according to Equation \ref{eq:SSEDist}: \begin{equation}
\label{eq:SSEDist}
m_3 = \sum^{K}_{k=1}(f(t_k) - \hat{f}_n(t_k))^2 \times d_k
\end{equation}

\item $m_4$ - \textit{Minimum Squared Error}: the minimum value of squared errors is calculated using the inverse of the distances $d_k$ as weights. The measure $m_4$ is computed using Equation \ref{eq:minSEDist}: \begin{equation}
\label{eq:minSEDist}
m_4 = \min_{1 \le k \le K}\{(f(t_k) - \hat{f}_n(t_k))^2 \times d_k\}
\end{equation}

\begin{table*}[!t]
\caption {Summary of the Competence Measures.} 
\centering
\label{tab:measures} 
\renewcommand{\arraystretch}{1.5} 
\begin{tabular}{|l|c|c|}
\hline
\multicolumn{1}{|c|}{\textbf{Measure}} & \multicolumn{1}{c|}{\textbf{Acronym}} & \multicolumn{1}{c|}{\textbf{Equation}} \\ \hline
Variance & $m_1$ & $Var({\hat{f}_n(t_1),\hat{f}_n(t_2),...,\hat{f}_n(t_K)})$ \\ \hline
Sum Absolute Error & $m_2$ & $\sum_{k=1}^{K}\left|f(t_k) - \hat{f}_n(t_k)\right| \times d_k$ \\ \hline
Sum Squared Error & $m_3$ & $\sum^{K}_{k=1}(f(t_k) - \hat{f}_n(t_k))^2 \times d_k$ \\ \hline
Minimum Squared Error & $m_4$ & $\min_{1 \le k \le K}\{(f(t_k) - \hat{f}_n(t_k))^2 \times d_k\}$ \\ \hline
Maximum Squared Error & $m_5$ & $\max_{1 \le k \le K}\{(f(t_k) - \hat{f}_n(t_k))^2 \times d_k\}$ \\ \hline
Neighbor's Similarity & $m_6$ & $\sum_{k=1}^{K}(f(t_k) - \hat{f}_n(x_j))^2 \times d_k$ \\ \hline
Root Sum Squared Error & $m_7$ & $\sum^{K}_{k=1} \sqrt{(f(t_k) - \hat{f}_n(t_k))^2 \times d_k}$ \\ \hline
Closest Squared Error & $m_8$ & $(f(t_1) - \hat{f}_n(t_1))^2$ \\ \hline
\end{tabular}
\end{table*}

\item $m_5$ - \textit{Maximum Squared Error}: the maximum value of squared errors is calculated using the inverse of the distances $d_k$ as weights. The measure $m_5$ is computed using Equation \ref{eq:maxSEDist}: \begin{equation}
\label{eq:maxSEDist}
m_5 = \max_{1 \le k \le K}\{(f(t_k) - \hat{f}_n(t_k))^2 \times d_k\}
\end{equation}

Considering that $m_4$ and $m_5$ define an interval, these measures present mean and variance, it means that, the interval contains information about implicit measures of dispersion (error variance) and centrality (error mean) of the squared error in the region of competence.

\item $m_6$ - \textit{Neighbor's Similarity}: the sum of the differences between the estimated values of the test pattern from test set $\mathcal{X}$ and the observed value of each neighborhood pattern, weighted by the inverse of the distance. The measure $m_6$ is computed using Equation \ref{eq:difRealValueDist}: \begin{equation}
\label{eq:difRealValueDist}
m_6 = \sum_{k=1}^{K}(f(t_k) - \hat{f}_n(x_j))^2 \times d_k
\end{equation} where $\hat{f}_n(x_j)$ is the estimated value by the regressor $\hat{f}_n$ for $x_j$.

The goal of the competence measure $m_6$ is to find the degree of similarity between the estimation of the pattern $x_j$ and the observed values of the nearest neighbors. This is the only measure that uses in its calculation the estimated value for the test pattern ($\hat{f}_n(x_j)$). So far as we know, this measure is unprecedented and is defined by the authors of this work.

\item $m_7$ - \textit{Root Sum Squared Error}: the root sum squared errors in region of competence, weighted by $d_k$. The measure $m_7$ is computed using Equation \ref{eq:RSSEDist}: \begin{equation}
\label{eq:RSSEDist}
m_7 = \sum^{K}_{k=1} \sqrt{(f(t_k) - \hat{f}_n(t_k))^2 \times d_k}
\end{equation}

The root squared error is more stable and less sensitivity to the difference between the maximum and the minimum errors, while the squared error is very sensitive to extreme error values. The measures $m_3$ and $m_7$ present different points of view from the error calculated in the region of competence. These two measures produce the same result when a single regressor is chosen to estimate a test pattern, but different results in the combination of the regressors.

\item $m_8$ - \textit{Closest Squared Error}: the error obtained by the regressor only on the nearest neighbor. The measure $m_8$ is computed using Equation \ref{eq:closeSE}: \begin{equation}
\label{eq:closeSE}
m_8 = (f(t_1) - \hat{f}_n(t_1))^2
\end{equation}

\end{itemize}


\section{Experiments}
\label{sec:experimentos}

\subsection{Datasets}

To show the performance of the DRS algorithms, a total of 15 regression datasets were used in the comparative study. The main features of each dataset are shown in Table \ref{tab:datasets}. These are public datasets, which are available in the following repositories: personal page of Prof. Luís Torgo\footnote{http://www.dcc.fc.up.pt/$\sim$ltorgo/Regression/DataSets.html}, UCI machine learning repository\footnote{http://archive.ics.uci.edu/ml/}, and Delve repository\footnote{http://www.cs.toronto.edu/$\sim$delve/}.

\begin{table}[!t]
\caption {Datasets used in the experiments.} 
\label{tab:datasets} 
\centering
\begin{tabular}{lccc}
\hline 
\textbf{Dataset} & \textbf{Instances} & \textbf{Features} & \textbf{Source} \\
\hline 
\textbf{Airfoil Self Noise} &  1503 & 5 & UCI\\
\textbf{Bank32NH} & 8192 & 32 & Delve\\
\textbf{Bank8FM} & 8192 & 8 & Delve\\
\textbf{Breast Cancer} & 194 & 32 & Torgo\\
\textbf{CCPP} & 9568 & 4 & UCI\\ 
\textbf{Concrete} & 1030 & 8 & UCI\\
\textbf{Delta Ailerons} & 7129 & 6 & Torgo\\
\textbf{Delta Elevators} & 9517 & 6 & Torgo\\
\textbf{Housing} & 506 & 13 & UCI\\ 
\textbf{Kinematics} & 8192 & 8 & Delve\\ 
\textbf{Machine} & 209 & 6 & Torgo\\ 
\textbf{Puma32H} & 8192 & 32 & Delve\\
\textbf{Puma8NH} & 8192 & 8 & Delve\\ 
\textbf{Stock} & 950 & 9 & Torgo\\
\textbf{Triazines} & 186 & 60 & Torgo\\ 
\hline
\end{tabular}
\end{table}

\begin{table*}[!t]
\centering
\caption{Mean results of the $MSE$ over 30 replications obtained for the DS algorithm and Individual Regressor. The best results are in bold. Line ``Win/Tie/Loss'' shows the total of the results. Error values are in the scale $10^{-4}$.}
\label{tab:dsprop}
\resizebox{1.0\textwidth}{!}{\begin{tabular}{lc|cccccccc}
\cline{3-10}
\multicolumn{1}{c}{} &    & \multicolumn{8}{c}{\textbf{DS}}                                                                                                                                                                                                        \\ \hline
\textbf{Dataset}       & \multicolumn{1}{l}{\textbf{\begin{tabular}[c]{@{}l@{}}Individual \\ Regressor\end{tabular}}} &  \multicolumn{1}{|c}{$m_1$} & \multicolumn{1}{c}{$m_2$} & \multicolumn{1}{c}{$m_3$} & \multicolumn{1}{c}{$m_4$} & \multicolumn{1}{c}{$m_5$} & \multicolumn{1}{c}{$m_6$} & \multicolumn{1}{c}{$m_7$} & \multicolumn{1}{c}{$m_8$} \\ \hline
Airfoil Self Noise&6.17(0.25)&10.47(0.36)&\textbf{4.58(0.21)}&4.66(0.18)&7.34(0.40)&4.70(0.22)&6.65(0.10)&4.66(0.18)&5.99(0.34)\\ 
Bank32NH&20.35(0.36)&21.16(0.37)&21.21(0.42)&21.14(0.34)&21.13(0.39)&21.14(0.38)&\textbf{19.51(0.06)}&21.14(0.34)&21.14(0.32)\\ 
Bank8FM&\textbf{2.69(0.04)}&3.05(0.06)&3.02(0.06)&3.01(0.05)&3.02(0.05)&3.01(0.05)&7.30(0.04)&3.01(0.05)&3.02(0.05)\\ 
Breast Cancer&122.17(7.29)&107.29(9.19)&128.44(9.23)&128.53(10.81)&129.84(9.34)&128.00(9.71)&\textbf{68.36(1.13)}&128.53(10.81)&126.62(10.12)\\ 
CCPP&3.04(0.06)&3.32(0.06)&3.30(0.06)&3.28(0.08)&3.75(0.08)&3.27(0.08)&\textbf{2.30(0.01)}&3.28(0.08)&3.69(0.07)\\ 
Concrete&6.26(0.27)&10.30(0.67)&6.23(0.40)&6.23(0.40)&8.60(0.58)&\textbf{6.21(0.48)}&9.74(0.23)&6.23(0.40)&7.65(0.47)\\ 
Delta Ailerons&2.27(0.05)&2.15(0.04)&2.53(0.05)&2.53(0.05)&2.53(0.06)&2.53(0.06)&\textbf{1.57(0.01)}&2.53(0.05)&2.52(0.05)\\ 
Delta Elevators&4.50(0.05)&4.25(0.05)&5.14(0.08)&5.13(0.07)&5.01(0.08)&5.10(0.07)&\textbf{3.04(0.01)}&5.13(0.07)&5.06(0.08)\\ 
Housing&9.73(0.95)&13.61(1.67)&9.95(1.22)&\textbf{9.71(1.48)}&11.10(1.34)&9.99(1.43)&10.49(0.55)&\textbf{9.71(1.48)}&10.84(1.36)\\ 
Kinematics&20.52(0.40)&19.94(0.37)&21.04(0.37)&21.01(0.36)&23.84(0.35)&20.96(0.36)&\textbf{6.41(0.03)}&21.01(0.36)&23.88(0.36)\\ 
Machine&\textbf{3.65(1.05)}&8.31(0.84)&4.15(1.53)&4.04(1.25)&5.11(1.89)&3.87(0.94)&4.35(0.68)&4.04(1.25)&4.23(1.40)\\ 
Puma32H&\textbf{3.54(0.03)}&3.91(0.05)&3.86(0.06)&3.86(0.05)&3.91(0.06)&3.87(0.05)&8.98(0.05)&3.86(0.05)&3.89(0.05)\\ 
Puma8NH&30.01(0.35)&31.13(0.46)&32.39(0.40)&32.40(0.39)&32.38(0.42)&32.36(0.52)&\textbf{22.94(0.09)}&32.40(0.39)&32.51(0.53)\\ 
Stock&1.48(0.14)&1.76(0.17)&1.41(0.16)&1.38(0.17)&1.89(0.21)&1.38(0.15)&\textbf{0.73(0.01)}&1.38(0.17)&1.76(0.18)\\ 
Triazines&\textbf{32.85(3.39)}&37.61(3.15)&36.94(4.84)&37.71(4.71)&39.56(4.30)&36.94(4.35)&32.92(0.95)&37.71(4.71)&38.60(5.21)\\  \hline
Win/Tie/Loss           & 4/0/11  & 0/0/15   &  1/0/14  &  0/1/14  &  0/0/14  &   1/0/14  &  8/0/7   &  1/0/14   &  0/0/15   \\ \hline
\end{tabular}}
\end{table*}

\begin{table*}[!t]
\centering
\caption{Mean results of the $MSE$ over 30 replications obtained for the DW algorithm, Mean and Median. The best results are in bold. Line ``Win/Tie/Loss'' shows the total of the results. Error values are in the scale $10^{-4}$.}
\label{tab:dwprop}
\resizebox{1.0\textwidth}{!}{\begin{tabular}{lcc|ccccccccc}
\cline{4-11}
\multicolumn{2}{c}{} &    & \multicolumn{8}{|c}{\textbf{DW}}                                                                                                                                                                                                        \\ \hline
\textbf{Dataset}       & \multicolumn{1}{c}{\textbf{Mean}} & \multicolumn{1}{c|}{\textbf{Median}} & \multicolumn{1}{c}{$m_1$} & \multicolumn{1}{c}{$m_2$} & \multicolumn{1}{c}{$m_3$} & \multicolumn{1}{c}{$m_4$} & \multicolumn{1}{c}{$m_5$} & \multicolumn{1}{c}{$m_6$} & \multicolumn{1}{c}{$m_7$} & \multicolumn{1}{c}{$m_8$} \\ \hline 
Airfoil Self Noise&2.84(0.06)&3.12(0.08)&3.96(0.17)&2.46(0.07)&2.31(0.08)&2.92(0.09)&\textbf{2.27(0.09)}&3.12(0.06)&2.50(0.07)&4.15(0.25)\\ 
Bank32NH&\textbf{10.81(0.06)}&11.26(0.08)&10.94(0.07)&10.87(0.06)&11.39(0.09)&11.15(0.07)&11.54(0.10)&13.34(0.05)&10.88(0.06)&15.35(0.18)\\ 
Bank8FM&\textbf{1.52(0.01)}&1.59(0.01)&\textbf{1.52(0.01)}&\textbf{1.52(0.01)}&1.55(0.01)&1.57(0.01)&1.56(0.01)&1.62(0.01)&\textbf{1.52(0.01)}&2.18(0.03)\\ 
Breast Cancer&72.58(1.82)&79.52(2.97)&71.99(1.88)&72.82(2.06)&74.26(2.59)&73.89(2.48)&74.73(2.71)&\textbf{70.82(1.69)}&72.84(2.01)&91.01(5.65)\\ 
CCPP&1.92(0.01)&\textbf{1.85(0.01)}&1.96(0.01)&\textbf{1.85(0.02)}&\textbf{1.85(0.02)}&1.98(0.02)&\textbf{1.85(0.02)}&1.87(0.01)&\textbf{1.85(0.02)}&2.62(0.04)\\ 
Concrete&3.91(0.13)&3.96(0.15)&4.19(0.13)&3.48(0.15)&\textbf{3.44(0.17)}&3.82(0.13)&\textbf{3.44(0.17)}&3.97(0.14)&3.48(0.14)&5.11(0.31)\\ 
Delta Ailerons&1.43(0.01)&1.48(0.01)&\textbf{1.42(0.01)}&1.45(0.01)&1.49(0.02)&1.45(0.01)&1.49(0.02)&1.43(0.01)&1.45(0.01)&1.58(0.02)\\ 
Delta Elevators&2.92(0.01)&3.02(0.01)&\textbf{2.89(0.01)}&2.96(0.01)&3.02(0.02)&2.93(0.01)&3.03(0.01)&\textbf{2.89(0.01)}&2.95(0.01)&3.18(0.02)\\ 
Housing&5.48(0.28)&5.48(0.55)&5.81(0.33)&5.21(0.31)&\textbf{5.17(0.36)}&5.51(0.26)&5.20(0.38)&6.01(0.30)&5.20(0.31)&6.98(0.71)\\ 
Kinematics&9.89(0.05)&9.46(0.06)&9.65(0.05)&9.42(0.05)&9.21(0.07)&10.33(0.06)&9.27(0.07)&\textbf{6.79(0.02)}&9.41(0.05)&15.32(0.19)\\ 
Machine&2.73(0.70)&2.86(0.98)&5.04(0.88)&\textbf{2.53(0.80)}&2.69(0.83)&2.88(0.64)&2.78(0.84)&3.26(0.81)&2.54(0.81)&3.42(0.98)\\ 
Puma32H&\textbf{1.94(0.01)}&1.98(0.01)&\textbf{1.94(0.01)}&1.95(0.01)&1.98(0.01)&2.04(0.01)&2.01(0.01)&2.05(0.01)&1.95(0.01)&3.04(0.04)\\ 
Puma8NH&\textbf{17.96(0.06)}&18.78(0.08)&17.97(0.06)&18.04(0.07)&18.36(0.10)&18.30(0.08)&18.48(0.11)&18.28(0.06)&18.05(0.07)&23.76(0.29)\\ 
Stock&0.87(0.04)&0.87(0.06)&0.95(0.08)&0.78(0.05)&0.75(0.05)&0.86(0.04)&0.74(0.05)&\textbf{0.62(0.03)}&0.77(0.05)&1.05(0.08)\\ 
Triazines&\textbf{23.59(1.35)}&25.40(2.08)&26.97(1.51)&23.91(1.53)&24.69(1.77)&24.12(1.45)&25.04(1.79)&25.97(1.26)&23.96(1.53)&29.89(3.04)\\   \hline
Win/Tie/Loss & 3/2/10 & 0/1/14 & 1/3/11 & 1/2/12 & 1/2/12 & 0/0/15 & 1/2/12 & 3/1/11 & 0/2/13 &  0/0/15\\  \hline
\end{tabular}}
\end{table*}

\begin{table*}[!t]
\centering
\caption{Mean results of the $MSE$ over 30 replications obtained for the DWS algorithm, Mean and Median. The best results are in bold. Line ``Win/Tie/Loss'' shows the total of the results. Error values are in the scale $10^{-4}$.}
\label{tab:dwsprop}
\resizebox{1.0\textwidth}{!}{\begin{tabular}{lcc|ccccccccc}
\cline{4-11}
\multicolumn{2}{c}{} &    & \multicolumn{8}{c}{\textbf{DWS}}                                                                                                                                                                                                        \\ \hline
\textbf{Dataset}       & \multicolumn{1}{c}{\textbf{Mean}} & \multicolumn{1}{c|}{\textbf{Median}} & \multicolumn{1}{c}{$m_1$} & \multicolumn{1}{c}{$m_2$} & \multicolumn{1}{c}{$m_3$} & \multicolumn{1}{c}{$m_4$} & \multicolumn{1}{c}{$m_5$} & \multicolumn{1}{c}{$m_6$} & \multicolumn{1}{c}{$m_7$} & \multicolumn{1}{c}{$m_8$} \\ \hline
Airfoil Self Noise&2.84(0.06)&3.12(0.08)&4.98(0.16)&\textbf{2.25(0.08)}&2.29(0.09)&2.92(0.09)&\textbf{2.25(0.09)}&4.01(0.08)&2.29(0.08)&4.15(0.25)\\ 
Bank32NH&\textbf{10.81(0.06)}&11.26(0.08)&11.06(0.08)&10.93(0.06)&11.41(0.09)&11.15(0.07)&11.55(0.10)&13.98(0.05)&10.95(0.06)&15.35(0.18)\\ 
Bank8FM&\textbf{1.52(0.01)}&1.59(0.01)&1.64(0.02)&1.54(0.01)&1.55(0.01)&1.57(0.01)&1.56(0.01)&3.03(0.03)&1.54(0.01)&2.18(0.03)\\ 
Breast Cancer&72.58(1.82)&79.52(2.97)&72.91(2.33)&74.37(2.67)&74.58(2.67)&73.89(2.48)&75.29(2.78)&\textbf{70.26(1.51)}&75.26(2.52)&91.01(5.65)\\ 
CCPP&1.92(0.01)&1.85(0.01)&2.08(0.01)&\textbf{1.83(0.02)}&1.84(0.02)&1.98(0.02)&1.85(0.02)&1.99(0.01)&\textbf{1.83(0.02)}&2.62(0.04)\\ 
Concrete&3.91(0.13)&3.96(0.15)&5.23(0.21)&\textbf{3.43(0.15)}&\textbf{3.43(0.17)}&3.82(0.13)&3.44(0.17)&5.33(0.15)&\textbf{3.43(0.14)}&5.11(0.31)\\ 
Delta Ailerons&\textbf{1.43(0.01)}&1.48(0.01)&\textbf{1.43(0.01)}&1.50(0.02)&1.50(0.02)&1.45(0.01)&1.50(0.02)&1.46(0.01)&1.51(0.02)&1.58(0.02)\\ 
Delta Elevators&2.92(0.01)&3.02(0.01)&\textbf{2.89(0.01)}&3.06(0.02)&3.04(0.02)&2.93(0.01)&3.04(0.02)&2.91(0.01)&3.09(0.02)&3.18(0.02)\\ 
Housing&5.48(0.28)&5.48(0.55)&7.42(0.43)&5.20(0.31)&\textbf{5.17(0.37)}&5.51(0.26)&5.20(0.38)&7.06(0.55)&5.21(0.32)&6.98(0.71)\\ 
Kinematics&9.89(0.05)&9.46(0.06)&9.75(0.06)&9.05(0.06)&9.16(0.07)&10.33(0.06)&9.23(0.07)&\textbf{6.50(0.02)}&9.01(0.07)&15.32(0.19)\\ 
Machine&2.73(0.70)&2.86(0.98)&6.15(0.78)&2.89(0.93)&\textbf{2.69(0.85)}&2.88(0.64)&2.78(0.85)&3.69(0.84)&2.97(0.96)&3.42(0.98)\\ 
Puma32H&\textbf{1.94(0.01)}&1.98(0.01)&2.04(0.01)&1.97(0.01)&1.99(0.01)&2.04(0.01)&2.01(0.01)&3.94(0.04)&1.98(0.01)&3.04(0.04)\\ 
Puma8NH&\textbf{17.96(0.06)}&18.78(0.08)&19.27(0.15)&18.37(0.09)&18.42(0.10)&18.30(0.08)&18.53(0.11)&19.70(0.07)&18.45(0.11)&23.76(0.29)\\ 
Stock&0.87(0.04)&0.87(0.06)&0.97(0.08)&0.75(0.05)&0.74(0.05)&0.86(0.04)&0.74(0.05)&\textbf{0.63(0.02)}&0.74(0.05)&1.05(0.08)\\ 
Triazines&\textbf{23.59(1.35)}&25.40(2.08)&27.50(1.58)&25.26(1.96)&24.99(1.84)&24.12(1.45)&25.30(1.83)&28.45(1.13)&25.53(1.96)&29.89(3.04)\\\hline
Win/Tie/Loss & 5/1/19 & 0/0/15 & 1/1/13 & 1/2/12 & 2/1/12 & 0/0/15 & 0/1/14 & 3/0/12 & 0/2/13 & 0/0/15 \\  \hline
\end{tabular}}
\end{table*}

\subsection{Experimental Protocol}
\label{sec:configExperimentos}

For each dataset, all data attributes are normalized into the interval [0,1], and the experiments were conducted using 30 replications. For each replication, the data is randomly split into ten parts of approximately the same size. Then, a 10-fold cross-validation is carried out using 90\% of the folds as the training set ($\mathcal{T}$) and 10\% as the testing set ($\mathcal{X}$).

In the experiments, homogeneous ensembles of size $N = 100$ were generated using Bagging~\cite{bagging}. Bagging generates different datasets, using sampling with replacement. Each generated dataset has the same size of the training set. Using replacement, some instances will be repeated in each dataset, and on average only 63\% of instances will be unique. The learning algorithm CART~\cite{Breiman1984} was used with default settings found in MATLAB without any specific adjustment.

In~\cite{Mendes-Moreira2009}, experiments were performed varying the size of the region of competence $K$ in the interval $\{2,4,6,8,10,12,14,16,18,20,25,30\}$. They concluded that the appropriate size for the neighborhood is problem-dependent, so they fixed the size of the region of competence with $K = 10$. Analyzing works of classification~\cite{Cruz2015a,Cruz2015}, time-series forecasting~\cite{Sergio2016}, and regression~\cite{Rooney2004}, it can be verified that the size of the region of competence is fixed for better validation and comparison of the results. So, all the experiments in this sections use $K = 10$ as the size of the region of competence.

For each test set, the Mean Squared Error - $MSE$ is computed. The result shows the arithmetic mean of the $MSE$ calculated for the 10 test sets used in the cross-validation. A single individual regressor was trained with the entire training set without the use of Bagging. The performance of this regressor is compared with the dynamic selection algorithm (DS), as will be presented in Section~\ref{sec:DSxPROP}.

\subsection{DS Results}
\label{sec:DSxPROP}

This section presents the results of the experiments performed using the DS algorithm. The experiments aim to compare the results obtained by the DS using each competence measure described in Section \ref{sec:caracteristicas}. Table \ref{tab:dsprop} shows the arithmetic mean of the results calculated over 30 replications for each dataset. Individual Regressor represents a single model trained using the whole training set $\mathcal{T}$, as explained in Section \ref{sec:configExperimentos}.

According to Table \ref{tab:dsprop}, the DS algorithm was better in 11 out of 15 and the individual regressor was better in only 4 out of 15 datasets. So, DRS is a good way to predict new test patterns, instead of a single regressor. Second, measure $m_6$ achieved the best performance. As pointed out earlier, only this measure uses the estimated value of the test pattern in its calculation. With the use of an ensemble with many regressors, this competence measure is interesting when a single regressor is selected from the ensemble.

Figure \ref{fig:DSGraph} shows the difference of the errors calculated in Table \ref{tab:dsprop} between measures $m_7$ and the best measure ($m*$) for each dataset. The difference of the errors in the datasets Bank8FM, Concrete, Housing, and Puma32H are zero or close to zero. We conclude that $m_7$ measure is not better in any dataset when DS algorithm is used.


\subsection{DW and DWS Results}
\label{sec:DSxPROP2}

Tables \ref{tab:dwprop} and \ref{tab:dwsprop} show the results to DW and DWS algorithms, respectively. Both algorithms combine the regressors from the ensemble. DW combine all the regressors using weighted mean and DWS selects and combine a subset of them.

Analyzing the results in Table \ref{tab:dwprop}, DW algorithm achieved better performance when compared with Mean and Median. Mean reached better performance in 3 out of 15 datasets, and Median has only one tie at dataset CCPP. Among the eight competence measures tested, six of them emerge with better results in DW, that is, none of them has superior performance for all datasets.

\begin{figure*}
\centering
\subfigure[DS algorithm]{%
\label{fig:DSGraph}%
\includegraphics[height=3in,width=3.3in]{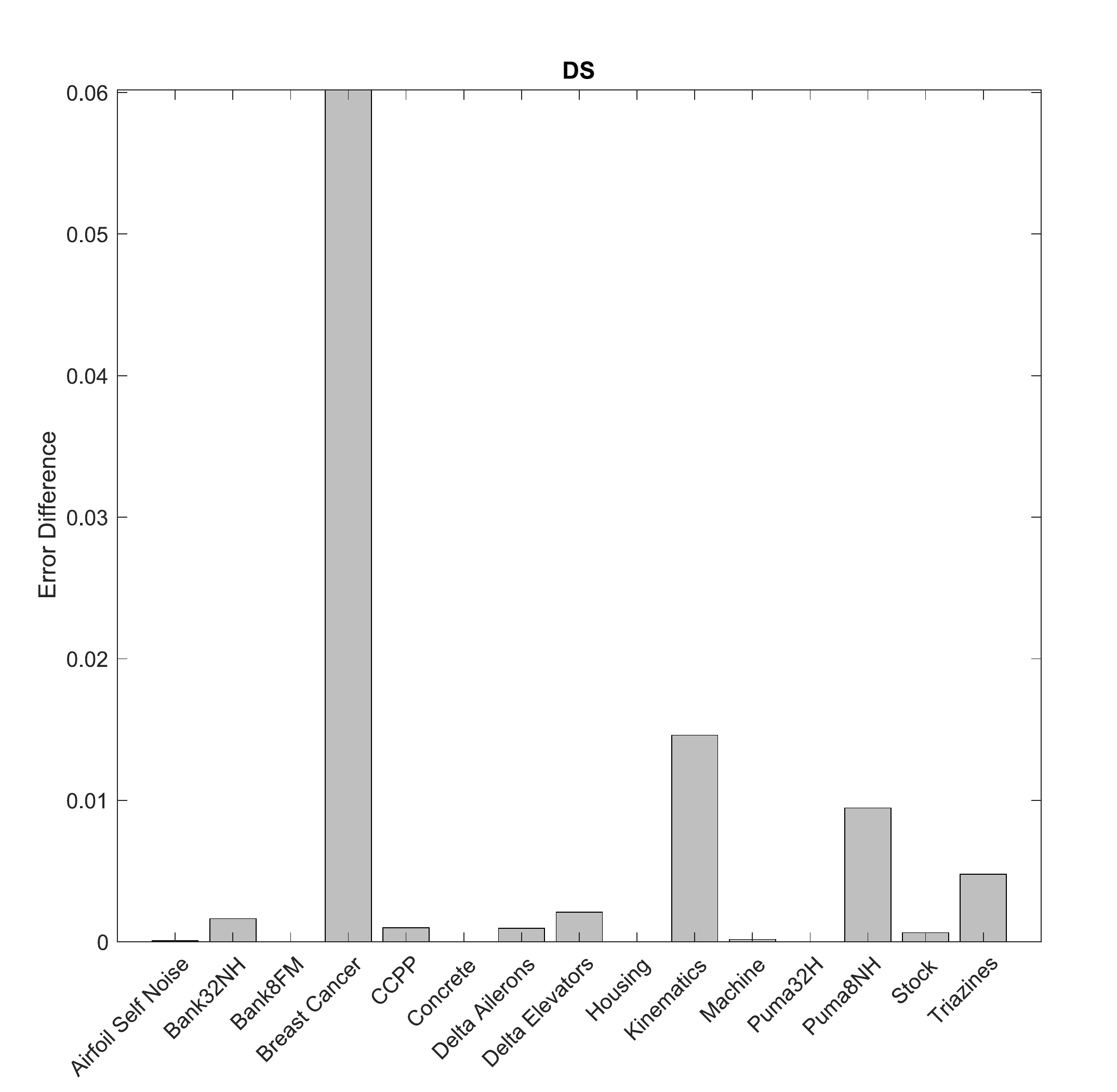}}
\qquad
\subfigure[DW algorithm]{%
\label{fig:DWGraph}%
\includegraphics[height=3in,width=3.3in]{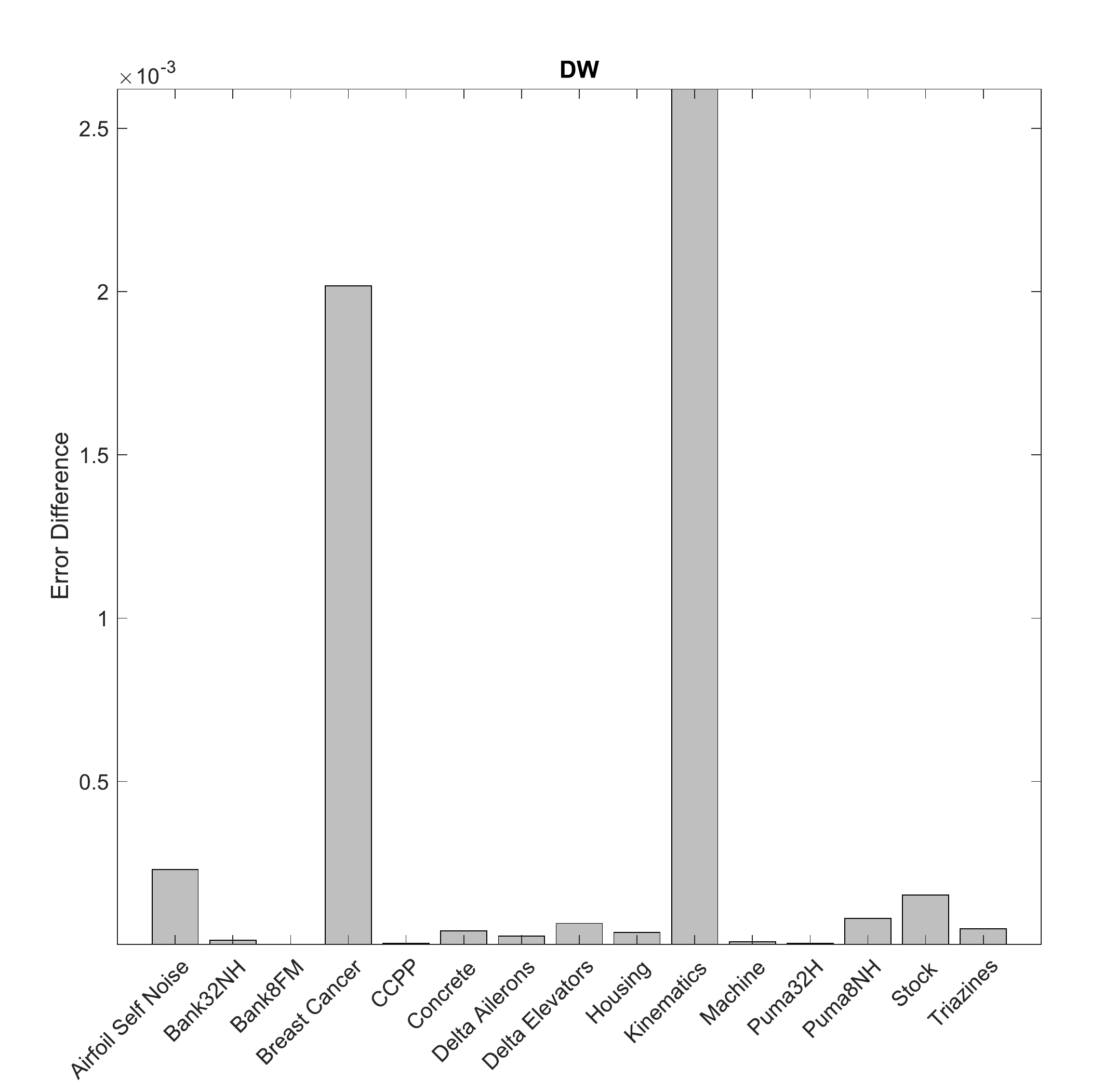}}
\qquad
\subfigure[DWS algorithm]{
\label{fig:DWSGraph}
\includegraphics[height=3in,width=3.3in]{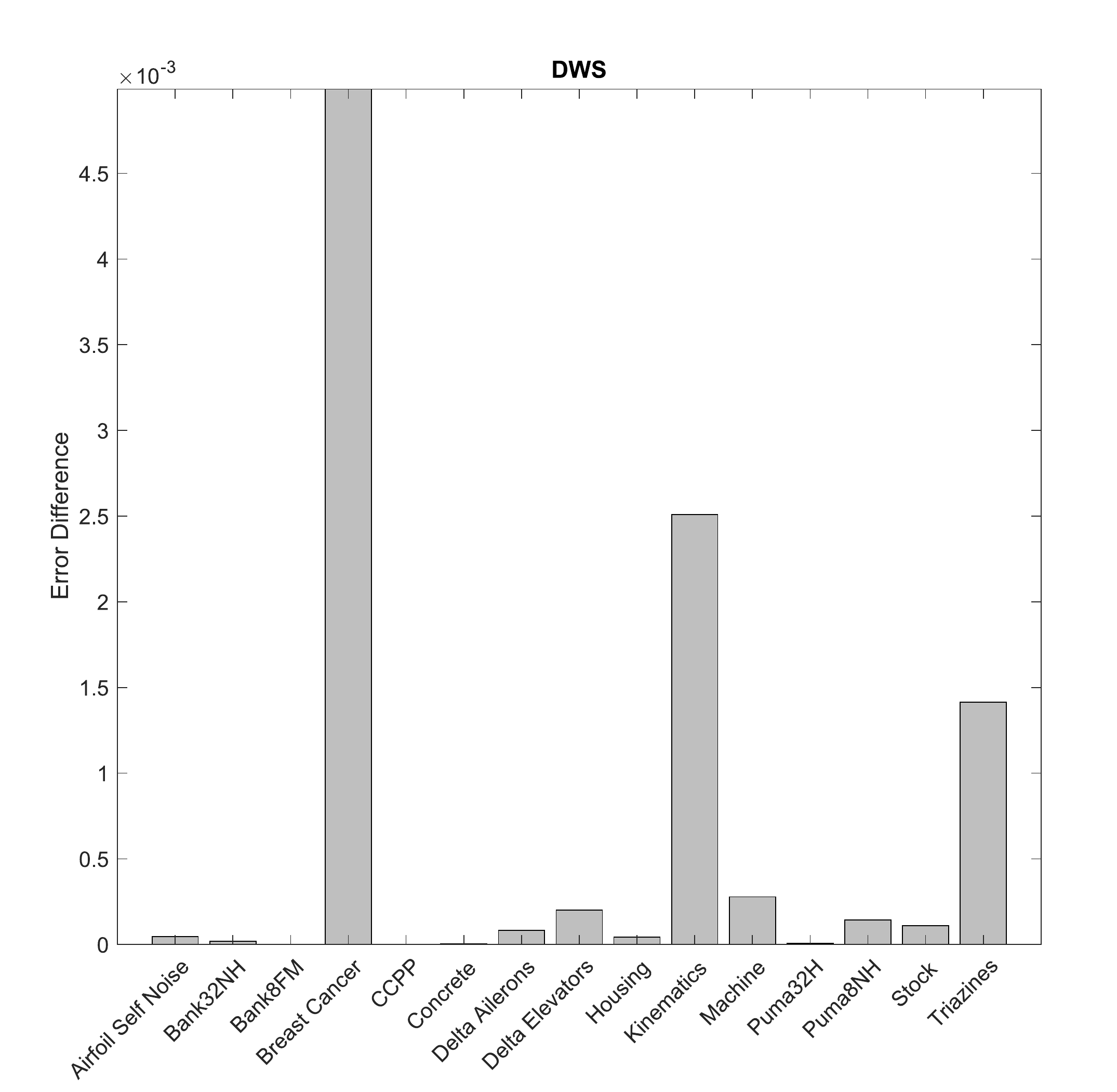}}
\caption{Comparison between measure $m_7$ and the best measure for each dataset. The bars present the difference of the errors between $m_7$ and $m*$, where $m*$ is the lowest error rate among the other measures.}
\end{figure*}

Figure \ref{fig:DWGraph} shows the difference of the errors calculated in Table \ref{tab:dwprop} between $m_7$ and the best measure ($m*$) for each dataset. Using the measures as weights for the combination of the regressors from the ensemble, as is done in DW algorithm, we conclude that $m_7$ is not better in any database when used with the DW algorithm. For the datasets Bank8FM, CCPP, and Puma32H, the difference of the errors are zero or close to zero.

In Table \ref{tab:dwsprop}, the DWS algorithm has better performance when compared with Mean and Median. Mean has better performance in 5 out of 15 datasets, and Median does not perform better in any dataset. Among the eight competence measures tested, six of them emerge with better results in DWS, that is, none of them has superior performance for all datasets.

Figure \ref{fig:DWSGraph} show the difference of the errors calculated in Table \ref{tab:dwsprop} between $m_7$ and the best measure. ($m*$) for each dataset. The same behavior observed with DW algorithm can be noticed here. $m_7$ is not better in any dataset and for the datasets Bank8FM, CCPP, Concrete, and Puma32H, the difference of the errors are zero or close to zero.

The experimental results show that $m_7$ proposed as competence measure in \cite{Mendes-Moreira2009} performs better in some datasets, but for others, there are competence measures that bring better overall performance for DRS systems.

\section{Conclusion}
\label{sec:conclusoes}

Differently from the literature on DCS, where several competence measures have been proposed and assessed during the last decade \cite{Britto2014,Cruz2018}, the number of works dealing with DRS is quite limited. The central issue in DRS, i.e., defining competence measures to help selecting the best regressor or ensemble of regressors, has been neglected in most of the works. To fill this gap, in this work we review eight competence measures, which were assessed using three different DRS systems and 15 datasets.

As presented in Section \ref{sec:experimentos}, DRS techniques perform better when compared to a single individual regressor or to classic statistical techniques such as Mean and Median. Another situation is that the reduction in the variance achieved by weighted average can explain why DW and DWS are better than DS~\cite{Tsymbal2006}.

It is possible to conclude that the competence measure used to select the regressors is problem-dependent. As previously mentioned, the literature techniques presented by~\cite{Mendes-Moreira2009}, Section \ref{sec:arquitetura}, use $m_7$ to select and combine the regressors, but our experiments pointed out that it does not have the best performance in all situations.

For future works, we can test a solution to select, for each regression problem, the best measure to be used in DRS techniques. Another solution is to combine the measures to select the most competent regressor or to use this combination as weighting to fuse the regressors from the ensemble.

\section*{Acknowledgment}

This research has been supported by the following Brazilian agencies: CNPq (Conselho Nacional de Desenvolvimento Científico e Tecnológico), CAPES (Coordenação de Aperfeiçoamento de Pessoal de Nível Superior) and FACEPE (Fundação de Amparo à Ciência e Tecnologia de Pernambuco).



%




\bibliographystyle{IEEEtran}
\bibliography{references}

\end{document}